\documentclass[conference]{IEEEtran}
\IEEEoverridecommandlockouts
\usepackage{amsmath,amssymb,amsfonts}
\usepackage{algorithmic}
\usepackage{graphicx}
\usepackage{textcomp}
\usepackage[authoryear,round]{natbib}
\usepackage[colorlinks=true, citecolor=black, linkcolor=black]{hyperref}  
\usepackage{multirow}
\usepackage{enumitem}
\usepackage[none]{hyphenat}
\def\BibTeX{{\rm B\kern-.05em{\sc i\kern-.025em b}\kern-.08em
    T\kern-.1667em\lower.7ex\hbox{E}\kern-.125emX}}
\usepackage{fancyhdr}
\usepackage{array}
\usepackage{booktabs}
\usepackage[table]{xcolor}
\hypersetup{
    pdftitle={An Evaluation of Deep Learning Models for Stock Market Trend Prediction},
    pdfauthor={Gonzalo López Gil, Paul Duhamel-Sebline, Andrew McCarren},
    pdfkeywords={Time Series, Prediction, Stock Market, xLSTM-TS, Wavelet Denoising, TCN, N-BEATS, TFT, N-HiTS, TiDE, S\&P 500, EWZ}
}
    
\begin{document}

\title{An Evaluation of Deep Learning Models for Stock Market Trend Prediction}

\author{
\IEEEauthorblockN{Gonzalo López Gil}
\IEEEauthorblockA{\textit{School of Computing} \\
\textit{Dublin City University}\\
Dublin, Ireland \\
gonzalo.lopezgil2@mail.dcu.ie}
\and
\IEEEauthorblockN{Paul Duhamel-Sebline}
\IEEEauthorblockA{\textit{School of Computing} \\
\textit{Dublin City University}\\
Dublin, Ireland \\
paul.duhamelsebline2@mail.dcu.ie}
\and
\IEEEauthorblockN{Andrew McCarren}
\IEEEauthorblockA{\textit{Insight Centre for Data Analytics} \\
\textit{Dublin City University}\\
Dublin, Ireland \\
andrew.mccarren@dcu.ie}
}

\maketitle

\begin{abstract}
The stock market is a fundamental component of financial systems, reflecting economic health, providing investment opportunities, and influencing global dynamics. Accurate stock market predictions can lead to significant gains and promote better investment decisions. However, predicting stock market trends is challenging due to their non-linear and stochastic nature.

This study investigates the efficacy of advanced deep learning models for short-term trend forecasting using daily and hourly closing prices from the S\&P 500 index and the Brazilian ETF EWZ. The models explored include Temporal Convolutional Networks (TCN), Neural Basis Expansion Analysis for Time Series Forecasting (N-BEATS), Temporal Fusion Transformers (TFT), Neural Hierarchical Interpolation for Time Series Forecasting (N-HiTS), and Time-series Dense Encoder (TiDE). Furthermore, we introduce the Extended Long Short-Term Memory for Time Series (xLSTM-TS) model, an xLSTM adaptation optimised for time series prediction.

Wavelet denoising techniques were applied to smooth the signal and reduce minor fluctuations, providing cleaner data as input for all approaches. Denoising significantly improved performance in predicting stock price direction. Among the models tested, xLSTM-TS consistently outperformed others. For example, it achieved a test accuracy of 72.82\% and an F1 score of 73.16\% on the EWZ daily dataset.

By leveraging advanced deep learning models and effective data preprocessing techniques, this research provides valuable insights into the application of machine learning for market movement forecasting, highlighting both the potential and the challenges involved.
\end{abstract}


\section{Introduction}
The stock market represents a complex system where prices are influenced by a myriad of factors, including economic indicators, market sentiment, geopolitical events, and corporate actions. Predicting stock market trends has long been an area of intense interest for researchers and experts. Successfully anticipating market movements can lead to significant financial returns and enhance the strategic decisions made by investors.

Traditional approaches to stock market prediction, such as time series analysis and econometric models, have provided some insights but are often limited by their linear assumptions and inability to handle large volumes of high-dimensional data. With the advent of big data and the exponential growth in computational power, deep learning has emerged as a promising tool for financial forecasting.

Deep learning models, with their ability to learn hierarchical representations and capture intricate patterns, offer a powerful alternative to conventional methods. In this study, the proposed xLSTM-TS model for predicting stock market trends is compared to several state-of-the-art models implemented using the Darts library, including TCN, N-BEATS, TFT, N-HiTS, and TiDE. Historical stock prices from the S\&P 500 index and the Brazilian ETF EWZ, previously denoised using wavelet techniques, are utilised to train and validate the models. Comprehensive experiments are conducted to evaluate these architectures and to analyse the results, understanding the advantages and limitations of each approach.

\section{Related Work}

This section reviews recent advancements in time series prediction, focusing on statistical models, data filtering and noise reduction, and deep learning techniques. Understanding these approaches and their performance provides a foundation for comparing different results and identifying the most effective methods.

\subsection{Statistical Models}

Statistical models, despite their simplicity, can often provide accurate predictions and sometimes outperform more complex techniques. ARIMA (AutoRegressive Integrated Moving Average) models are frequently used for predictive purposes. \cite{7046047} applied ARIMA to Nokia and Zenith Bank indices, demonstrating its competitiveness with emerging techniques for short-term predictions.

Recent studies have further explored the effectiveness of statistical methods in various applications. \cite{9498451} compared the performance of ARIMA with more advanced algorithms, such as LSTMs and Facebook's FBProphet, showing that while advanced methods can outperform ARIMA in some cases, ARIMA remains competitive for certain types of data. Additionally, \cite{9736353} conducted a comprehensive evaluation of different time series forecasting techniques, concluding that statistical models excel with datasets exhibiting low stochasticity.

However, statistical methods have limitations. For example, a study on Apple stock using SARIMA (Seasonal ARIMA) found it unsuitable due to the model's inability to handle non-linear stock market behaviour \citep{9609720}. Hybrid approaches that combine different statistical techniques have shown promise, but even these sometimes fail to capture complex patterns in financial data. These limitations highlight the need for more sophisticated methods to improve prediction accuracy in time series forecasting.

\subsection{Data Filtering and Noise Reduction}

Effective data filtering and noise reduction techniques are essential preprocessing steps for enhancing the accuracy of financial time series models. By cleaning the signal and reducing noise, these methods improve the quality of input data, leading to more reliable forecasts.

Wavelet transforms analyse stock market trends over different periods and often show superior performance. \cite{info12100388} demonstrated that combining multiresolution wavelet reconstruction with deep learning significantly improves medium-term stock prediction accuracy, achieving a 75\% hit rate for US stocks. Another study introduced the Adaptive Multi-Scale Wavelet Neural Network (AMSW-NN), which performs well but depends on dataset quality \citep{info12060252}.

Smoothing techniques like Simple Moving Average (SMA) and Exponential Moving Average (EMA) are commonly used to filter data and reveal long-term trends. \cite{9596479} examined the effectiveness of SMA and EMA for predicting Ultratech and Fortis shares. These traditional smoothing techniques often struggle with capturing the importance of diverse time steps, whereas the Self-Attentive Moving Average (SAMA) model incorporates an attention mechanism to overcome these limitations, showing better performance than traditional indicators \citep{app12073602}.

The Kalman filter, which estimates the state of a system from noisy measurements, has notable applications in finance. \cite{9144247} proposed a Possibilistic Kalman Filter to improve model robustness to uncertain data inputs, though its practical application in finance remains unexplored.

\subsection{Deep Learning}

Deep learning has made significant strides in time series forecasting by uncovering complex patterns in large datasets. Models such as Recurrent Neural Networks (RNNs), including Long Short-Term Memory (LSTM) networks, excel at analysing financial data and predicting market trends.

RNNs, particularly LSTMs, are adept at processing sequential data by maintaining information over extended periods. LSTMs have proven highly effective for financial market predictions, as demonstrated by \cite{7966019}, which showcases LSTM's ability to predict stock movements on the Brazilian Stock Exchange. Additionally, a study by \cite{Touzani2021} combines LSTM with Gated Recurrent Unit (GRU) models for short- and medium-term predictions, achieving remarkable accuracy for high-volume Moroccan stocks.

Hybrid models that integrate multiple deep learning architectures or combine them with traditional methods further enhance prediction accuracy. For instance, the hybrid model combining a bidirectional RNN with the African buffalo optimisation algorithm (ABBRM) achieved high prediction accuracies for Indian companies \cite{pattewar2023buffalo}.

Convolutional Neural Networks (CNNs), originally designed for image processing, have also been adapted for time series analysis in financial markets, allowing for the extraction of local temporal features and capturing long-term dependencies. The Local Graph Convolutional Network (LoGCN) introduced by \cite{WANG2023108687} employs graph-based convolution to forecast market trends, demonstrating superior performance on Chinese stock indices.

Overall, the application of deep learning techniques in financial forecasting offers significant advancements, enhancing the accuracy and robustness of predictive models and improving our understanding of market dynamics.

\section{Methodology}

\subsection{Dataset}

In this paper, the datasets were obtained from Yahoo Finance and Tiingo. The Yahoo Finance API, which provides historical data for a variety of financial markets and products, was utilised to obtain daily prices. For hourly price data, the Tiingo API was employed due to its fewer constraints. The datasets used in our research include the American index S\&P 500 and the Brazilian Exchange Traded Fund (ETF) EWZ.

The S\&P 500 is one of the most commonly used indices in the literature worldwide for the prediction of stock market prices and trends. This allows us to compare our results with previous studies and a wider array of models. In contrast, the Brazilian ETF EWZ replicates the performance of a group of mid-cap companies on Sao Paulo's Stock Exchange. The goal with this ETF is to evaluate our models on an index with higher volatility. It is also likely that the American market is more efficient due to the higher number of financial institutions actively monitoring and addressing irregularities, resulting in quicker detection of patterns and anomalies, and thus making it more challenging to predict.

For each of these indices, we retrieved both daily and hourly data. The daily datasets contain approximately 6000 rows, and the hourly datasets contain around 6300 rows.

\vspace{-0.1cm}
\begin{table}[htbp]
\centering
\caption{Start and End Dates of Datasets}
\begin{tabular}{|c|c|c|}
  \hline
  \textbf{Dataset} & \textbf{Start Date} & \textbf{End Date} \\
  \hline
  EWZ Daily & 14/07/2000 & 29/12/2023 \\
  \hline
  S\&P 500 Daily & 03/01/2000 & 29/12/2023 \\
  \hline
  EWZ Hourly & 13/072020 & 11/07/2024 \\
  \hline
  S\&P 500 hourly & 13/07/2020 & 11/07/2024 \\
  \hline
\end{tabular}
\label{tab:datasets}
\end{table}

The datasets provide information on Date, High, Low, Close, Adjusted Close, and Volume, as shown in Table \ref{tab-dataset}. For this study, we focused on the closing price (Close) as our target variable to predict the next day's trend.

\vspace{-0.1cm}
\begin{table}[htbp]
\centering
\caption{Dataset Sample from the S\&P 500}
\begin{tabular}{|c|c|c|c|c|c|}
\hline
\textbf{Date} & \textbf{Open} & \textbf{High} & \textbf{Low} & \textbf{Close} & \textbf{Volume} \\ \hline
03/01/2000 & 1469.25 & 1478.00 & 1438.36 & 1455.22 & 931800000 \\ \hline
04/01/2000 & 1455.22 & 1455.22 & 1397.43 & 1399.42 & 1009000000 \\ \hline
05/01/2000 & 1399.42 & 1413.27 & 1377.68 & 1402.11 & 1085500000 \\ \hline
06/01/2000 & 1402.11 & 1411.90 & 1392.10 & 1403.45 & 1092300000 \\ \hline
07/01/2000 & 1403.45 & 1441.47 & 1400.73 & 1441.47 & 1225200000 \\ \hline
\end{tabular}
\label{tab-dataset}
\end{table}

\subsection{Data Collection and Preprocessing} 

Preprocessing is a crucial step in preparing a dataset for machine learning tasks. This process involves several steps that clean, transform, and structure the data to ensure the model can learn effectively.

The initial step in preprocessing is denoising, which focuses on removing noise and irrelevant or erroneous data points that can obscure the patterns needed for effective model learning (see Section~\ref{subsec:noise-reduction}).

Next, converting dates and times to a correct and consistent format is essential when dealing with temporal data. We converted the timestamps to ``time-zone naive'' date-times, ensured there were no null values, and verified that our time index included exclusively business days.

Splitting the dataset into training, validation, and test sets is fundamental for evaluating the model's performance and ensuring it generalises well to unseen data. The training set is used to train the model, the validation set helps tune hyperparameters and prevent overfitting, and the test set is used for the final evaluation of the model's performance. The specific start and end dates, along with the percentages of each split, are detailed in Table \ref{tab:dataset-splits}.

\vspace{-0.1cm}
\begin{table}[htbp]
\centering
\caption{Start and End Dates for Dataset Splits}
\begin{tabular}{|c|c|c|c|c|}
  \hline
  \textbf{Dataset} & \textbf{Split} & \textbf{Start Date} & \textbf{End Date} & \textbf{Percentage} \\
  \hline
  \multirow{3}{*}{Daily} & Training   & 01/01/2000 & 31/12/2020 & 86\% \\
  & Validation & 01/01/2021    & 30/06/2022 & 7\% \\
  & Test       & 01/07/2022 & 31/12/2023 & 7\% \\
  \hline
  \multirow{3}{*}{Hourly} & Training & 13/07/2020 & 30/06/2023 & 75\% \\
  & Validation & 01/07/2023    & 31/12/2023 & 12,5\% \\
  & Test       & 01/01/2024 & 11/07/2024 & 12,5\% \\
  \hline
\end{tabular}
\label{tab:dataset-splits}
\end{table}

Other critical preprocessing steps include normalisation and sequencing. Normalisation scales the data to a standard range between 0 and 1, which helps speed up the training process and achieve better convergence. Sequencing is particularly important for time series data; this involves converting raw data into sequences of fixed-length windows, where each window serves as an input sequence and the subsequent value as the target. The length of each sequence is also an important factor for some models.

Finally, depending on the library used, we converted the dataset into the appropriate format. For instance, the Darts library \citep{JMLR:v23:21-1177}, a modern machine learning library for time series analysis, requires datasets to be in the TimeSeries format. This library provides a user-friendly interface and advanced functionalities for building and evaluating time series forecasting models.

\subsection{Noise Reduction}
\label{subsec:noise-reduction}

Noise in financial time series can significantly degrade the performance of prediction models. To mitigate this issue, we implemented wavelet denoising, inspired by \cite{1010023071}. This method was chosen for its powerful noise reduction and feature extraction capabilities.

Wavelet transform (WT) effectively decomposes non-stationary signals into approximation and detail coefficients, capturing both frequency and time information. We used the discrete wavelet transform (DWT) with the Daubechies wavelet (db4) for its compact support and orthogonality, based on its proven effectiveness in previous studies \citep{BOLZAN2020299, omidvar2021eeg}.

The denoised process involved padding the data to mitigate boundary effects, decomposing the signal using DWT, estimating noise levels, applying soft thresholding to detail coefficients, and reconstructing the denoised signal by setting high-frequency coefficients to zero. This approach effectively reduced noise while preserving essential features of the data.

\begin{figure}[htbp]
\centerline{\includegraphics[width=\columnwidth]{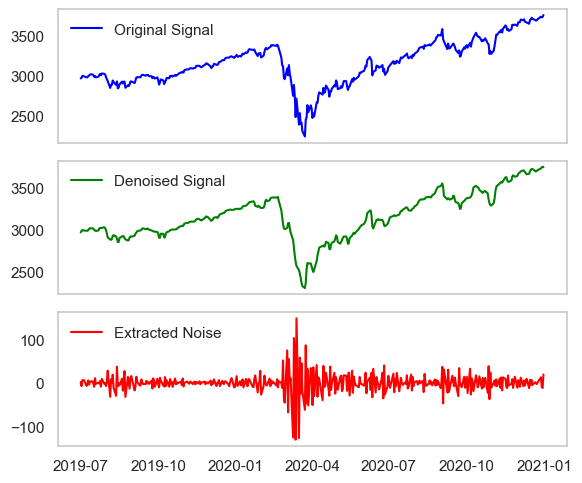}}
\caption{Wavelet denoising of S\&P 500 close prices during COVID-19.}
\label{fig-wavelet}
\end{figure}

Figure \ref{fig-wavelet} illustrates the decomposition process of DWT applied to the daily close prices of the S\&P 500. The original time series (top plot) contained significant noise, particularly noticeable during the COVID-19 period in 2020. The isolated noise component is shown in the bottom plot. The improved signal-to-noise ratio (SNR) in the denoised data underscores the success of this preprocessing step.

\subsection{Model Descriptions}

\vspace{0.1cm}
\begin{itemize}[label=1)]
\item{\textit{Proposed Model: xLSTM-TS}}
\end{itemize}
\vspace{0.1cm}

Traditional LSTM models have been widely successful in sequence forecasting tasks; however, they often encounter limitations such as gradient vanishing and inadequate handling of long-term dependencies, particularly in time series forecasting. These challenges necessitate the development of an improved architecture that can enhance model robustness.

In this study, we introduce the Extended Long Short-Term Memory for Time Series (xLSTM-TS) model, inspired by the xLSTM architecture proposed by \cite{beck2024xlstmextendedlongshortterm}. The xLSTM improves upon traditional LSTM through exponential gating and new memory structures.

Exponential gating provides robust and stabilised gating functions, enhancing the model's ability to manage long-term dependencies. The novel memory structures include sLSTM with scalar memory and mLSTM with matrix memory, which improve parallelisability and efficiency. Specifically, sLSTM simplifies memory management with scalar updates, while mLSTM supports full parallelisation by eliminating hidden-hidden recurrent connections.

Integrating these LSTM extensions into residual block backbones results in xLSTM blocks, which are residually stacked to form the xLSTM architecture. This design leverages the strengths of both sLSTM and mLSTM, offering a flexible and powerful framework for sequence modelling tasks.

Our xLSTM-TS model adapts the xLSTM for time series forecasting by using the \texttt{xLSTMBlockStack} and adjusting key parameters to optimise performance while managing computational constraints. Inspired by the official xLSTM implementation, we modified it for a time series-specific approach. The detailed architecture of the xLSTM-TS model is summarised in Table \ref{table_xlstm_summary}.

\begin{table}[htbp]
\caption{Summary of the xLSTM-TS Model Architecture}
\begin{center}
\begin{tabular}{|l|l|r|}
\hline
\textbf{Layer} & \textbf{Output Shape} & \textbf{Param \#} \\
\hline
xLSTM-TS Model & [16, 1] & -- \\
\hspace{0.3cm} Linear & [16, 150, 64] & 128 \\
\hspace{0.3cm} xLSTMBlockStack & [16, 150, 64] & -- \\
\hspace{0.6cm} ModuleList & -- & -- \\
\hspace{0.9cm} mLSTMBlock & [16, 150, 64] & 27,844 \\
\hspace{0.9cm} sLSTMBlock & [16, 150, 64] & 41,600 \\
\hspace{0.9cm} mLSTMBlock & [16, 150, 64] & 27,844 \\
\hspace{0.9cm} mLSTMBlock & [16, 150, 64] & 27,844 \\
\hspace{0.6cm} LayerNorm & [16, 150, 64] & 64 \\
\hspace{0.3cm} Linear & [16, 1] & 65 \\
\hline
\textbf{Total params:} & & \textbf{125,389} \\
\hline
\end{tabular}
\label{table_xlstm_summary}
\end{center}
\end{table}

Key configuration settings are listed in Table \ref{table_key_config}. The input size was set to 1, representing the number of features in the time series (in this case, the closing price). An embedding dimension of 64 was chosen to balance memory usage and performance. The output size was also set to 1, indicating that the model predicts the next value in the series (the next day or hour). A sequence length of 150 was used to capture sufficient historical data without incurring excessive computational cost. The batch size was set to 16 to avoid GPU memory limits. The context length was aligned with the sequence length to ensure consistency in data fed into the model.

\begin{table}[htbp]
\caption{Key Configuration Settings}
\begin{center}
\begin{tabular}{|c|c|}
\hline
\textbf{Parameter} & \textbf{Value} \\
\hline
Input Size & 1 \\
Embedding Dimension & 64 \\
Output Size & 1 \\
Sequence Length & 150 \\
Batch Size & 16 \\
Context Length & 150 \\
\hline
\end{tabular}
\label{table_key_config}
\end{center}
\end{table}

The configurations for the mLSTM and sLSTM blocks are detailed in Tables \ref{table_mlstm} and \ref{table_slstm}, respectively.

\begin{table}[htbp]
\caption{mLSTM Block Configuration for xLSTM-TS}
\begin{center}
\begin{tabular}{|c|c|}
\hline
\textbf{Parameter} & \textbf{Value} \\
\hline
Convolutional Kernel Size & 4 \\
Projection Block Size & 2 \\
Number of Heads & 2 \\
\hline
\end{tabular}
\label{table_mlstm}
\end{center}
\end{table}

\begin{table}[!htbp]
\caption{sLSTM Block Configuration for xLSTM-TS}
\begin{center}
\begin{tabular}{|c|c|}
\hline
\textbf{Parameter} & \textbf{Value} \\
\hline
Convolutional Kernel Size & 2 \\
Number of Heads & 2 \\
Feedforward Projection Factor & 1.1 \\
\hline
\end{tabular}
\label{table_slstm}
\end{center}
\end{table}

The xLSTM architecture was implemented with four blocks, each consisting of a combination of sLSTM and mLSTM layers. The model's ability to process long sequences efficiently was leveraged by modifying the context length and embedding dimension as mentioned before.

Additionally, linear layers were added to project the input data to the required embedding dimension and to project the xLSTM output to the desired output size. For instance, the input projection layer was defined as \texttt{nn.Linear(input\_size, embedding\_dim)}.

To manage the data, a custom DataLoader was created to batch the input time series data. This DataLoader ensured that the data was fed into the model in manageable chunks, maintaining the integrity and sequential nature of the time series data.

This xLSTM-TS setup provided a robust framework for time series forecasting, balancing computational efficiency with predictive accuracy. The modifications and configurations ensured that the model could handle large datasets effectively while maintaining high performance, making it a suitable choice for our predictive modelling tasks.

\vspace{0.1cm}
\begin{itemize}[label=2)]
\item{\textit{Existing Models (Darts Library)}}
\end{itemize}
\vspace{0.1cm}

The library we used to implement the state-of-the-art models is the Darts library, which contains a variety of models for series forecasting. As the following models come from the same library, they share a certain amount of parameters such as the Sequence Length, set to 100, and the Batch Size, set to 256. These values were chosen so that performance is optimised with a reasonable computation time.

\begin{table}[htbp]
\caption{Common parameters for Darts Models}
\begin{center}
\begin{tabular}{|c|c|}
\hline
\textbf{Parameter} & \textbf{Value} \\
\hline
Sequence Length & 100 \\
Batch Size & 256 \\
\hline
\end{tabular}
\label{table_darts_models}
\end{center}
\end{table}

\begin{itemize}[label=-]
\item{\textit{TCN}}
\end{itemize}

Advancements in sequence modelling have introduced Temporal Convolutional Networks (TCNs) as a powerful alternative to traditional recurrent neural networks (RNNs) like LSTMs and GRUs. TCNs leverage convolutional architectures to capture temporal dependencies within sequential data, offering several advantages over their recurrent counterparts \citep{bai2018empiricalevaluationgenericconvolutional}.

The TCN architecture employs causal convolutions, ensuring that predictions at any time step depend only on current and past inputs, preventing information leakage from future data points. A key feature of TCNs is the use of dilated convolutions, which allow the network to have a large receptive field without increasing depth excessively. This enables TCNs to model long-range temporal patterns effectively. Residual connections are also incorporated to facilitate the training of deep networks by mitigating gradient-related issues.

Concerning the parameters of our TCN, the Dropout was set to zero to allow a faster training there wasn't any overfitting issues with it. Also, as the model should capture long-term dependencies, without needing a deep network, and to capture more context from the input, the dilation base was set to 2 and the kernel size to 7. A balance also had to be found between computational efficiency and the ability to learn complex patterns, hence the four filters.

\begin{table}[htbp]
\caption{TCN Configuration}
\begin{center}
\begin{tabular}{|c|c|}
\hline
\textbf{Parameter} & \textbf{Value} \\
\hline
Dropout & 0 \\
Dilation Base & 2 \\
Kernel Size & 7 \\
Number Filters & 4 \\
\hline
\end{tabular}
\label{table_tcn_configuration}
\end{center}
\end{table}

\begin{itemize}[label=-]
\item{\textit{N-BEATS}}
\end{itemize}

The Neural Basis Expansion Analysis for Time Series Forecasting (N-BEATS) model uses a deep neural architecture with backward and forward residual connections to enhance gradient flow and training stability. The model processes a lookback window of historical data, generating both a forecast (forward prediction) and a backcast (reconstruction of the input). Residuals, the differences between the actual input and the backcast, are passed to subsequent blocks, progressively refining the predictions. This hierarchical approach allows the model to capture both short-term fluctuations and long-term trends effectively \citep{oreshkin2020nbeatsneuralbasisexpansion}.

N-BEATS employs backward residuals to capture unexplained variance and forward residuals to facilitate gradient flow. The final output is an aggregation of forecasts from each block, enhancing predictive accuracy.

To balance learning complex features with computational efficiency, the parameters outlined in Table \ref{table_nbeats_configuration} were chosen.

\begin{table}[htbp]
\caption{N-BEATS Configuration}
\begin{center}
\begin{tabular}{|c|c|}
\hline
\textbf{Parameter} & \textbf{Value} \\
\hline
Number of Stacks & 10 \\
Number of Blocks per Stack & 1 \\
Number of Layers per Block & 4 \\
Layer Widths & 512 \\
Epochs per Validation Period & 1 \\
\hline
\end{tabular}
\label{table_nbeats_configuration}
\end{center}
\end{table}

\begin{itemize}[label=-]
\item{\textit{TFT}}
\end{itemize}

The Temporal Fusion Transformer (TFT) is a significant advancement in multi-horizon time series forecasting, leveraging attention mechanisms to capture complex temporal patterns \citep{LIM20211748}. TFT integrates high-performance forecasting capabilities with enhanced interpretability, making it a state-of-the-art model for time series analysis.

The TFT architecture includes recurrent layers for local temporal dependencies, self-attention layers for long-term dependencies, and gating layers to eliminate unnecessary components. This combination allows TFT to improve performance while providing insights into temporal dynamics.

TFT processes static covariates, past observed inputs, and known future inputs through embedding layers. The LSTM layer captures short-term dependencies, with a hidden size of 64 ensuring sufficient capacity to learn intricate patterns. The multi-head attention mechanism with four heads enables the model to focus on multiple aspects of the input data simultaneously, capturing long-term dependencies and interactions between different time steps. Gating layers, including Gated Residual Networks (GRNs), control the flow of information and suppress irrelevant parts of the network, allowing the model to adjust its complexity based on the dataset. The key configuration parameters for the TFT model are outlined in Table \ref{table_tft_configuration}.

\begin{table}[htbp]
\caption{TFT Configuration}
\begin{center}
\begin{tabular}{|c|c|}
\hline
\textbf{Parameter} & \textbf{Value} \\
\hline
Hidden Size & 64 \\
LSTM Layers & 1 \\
Number of Attention Heads & 4 \\
Dropout & 0.1 \\
\hline
\end{tabular}
\label{table_tft_configuration}
\end{center}
\end{table}

\begin{itemize}[label=-]
\item{\textit{N-HiTS}}
\end{itemize}

The N-HiTS (Neural Hierarchical Interpolation for Time Series) model is a state-of-the-art approach designed to address challenges in long-horizon forecasting, such as prediction volatility and computational complexity \citep{challu2022nhitsneuralhierarchicalinterpolation}. This model improves both accuracy and efficiency through hierarchical interpolation and multi-rate data sampling.

N-HiTS employs multi-rate data sampling to capture different frequency components by subsampling the input data at various rates. Each block in the model processes data at a specific scale, reducing computational load while capturing long-term dependencies. Hierarchical interpolation constructs forecasts by combining the outputs from multiple blocks, ensuring smooth and consistent long-term predictions. Each block generates coefficients for both backcast and forecast outputs, which are interpolated to provide the final forecast. The block structure includes multiple stacks, each with several blocks comprising a multilayer perceptron (MLP) with ReLU activations, performing non-linear projections onto basis functions to generate the necessary coefficients.

\begin{itemize}[label=-]
\item{\textit{TiDE}}
\end{itemize}

The Time-series Dense Encoder (TiDE), introduced by \cite{das2024longtermforecastingtidetimeseries}, is a novel MLP-based encoder-decoder model for long-term time-series forecasting. TiDE combines the simplicity and computational efficiency of linear models with the capability to handle non-linear dependencies and covariates.

TiDE's architecture includes several key components. The Feature Projection step uses a residual block to project dynamic covariates into a lower-dimensional space. The Dense Encoder concatenates the look-back series, projected covariates, and static attributes, embedding them through a series of residual blocks. The Dense Decoder then transforms the encoded representation into future time-step vectors using additional residual blocks. The Temporal Decoder combines these vectors with projected covariates to generate final predictions. A global residual connection maps the look-back series to the horizon, ensuring purely linear models are a subclass of TiDE.

\section{Experimental Setup}

\subsection{Hyper-parameters Tuning}

\begin{itemize}[label=-]
\item{\textit{xLSTM-TS}}
\end{itemize}

For the xLSTM-TS model, hyper-parameter tuning was essential to achieve optimal performance for time series prediction. The key parameters included learning rate, number of epochs, batch size, patience for early stopping, and gradient clipping max norm.

The Adam optimiser was selected for training due to its efficiency in handling sparse gradients, which are common in time series data. The learning rate was initially set to 0.0001, after preliminary experiments indicated that this value allowed stable convergence without excessively long training times. A learning rate scheduler was employed to reduce the learning rate by a factor of 0.5 if the validation loss did not improve for 10 consecutive epochs, aiding in fine-tuning the model's performance.

Training was conducted for a maximum of 200 epochs. Early stopping was implemented to avoid overfitting, and configured with a patience parameter of 30, meaning training would stop if the validation loss did not improve for 30 consecutive epochs.

Batch size, crucial for memory management on the GPU, was set to 16. This size was determined to be a suitable compromise between memory usage and the stability of gradient estimates.

Gradient clipping was applied with a maximum norm of 1.0 to mitigate the exploding gradient problem, which can be particularly problematic in deep models like the LSTM-based ones.

\begin{table}[htbp]
\caption{Optimal Hyper-parameter Configuration for xLSTM-TS}
\begin{center}
\begin{tabular}{|c|c|}
\hline
\textbf{Parameter} & \textbf{Value} \\
\hline
Learning Rate & 0.0001 \\
Batch Size & 16 \\
Sequence Length & 150 \\
Number of Epochs & 200 \\
Patience & 30 \\
Gradient Clipping Max Norm & 1.0 \\
Optimiser & Adam \\
Loss Function & MSE \\
\hline
\end{tabular}
\label{table_hyperparameters_xlstm}
\end{center}
\end{table}

The tuning process involved a systematic grid search over a range of hyper-parameters, including variations in learning rate (0.0001, 0.0005, 0.001), batch size (8, 16, 32), and sequence length (100, 150, 200). Each configuration was evaluated based on validation loss to identify the optimal settings.

The final configuration, as detailed in Table \ref{table_hyperparameters_xlstm}, provided the best trade-off between computational efficiency and prediction accuracy. This thorough tuning process ensured that the xLSTM-TS model was well-suited for time series forecasting tasks, demonstrating robust performance across the chosen datasets.

\begin{itemize}[label=-]
\item{\textit{Existing Models (Darts Library)}}
\end{itemize}

For the darts models, the hyper-parameters used are mostly the same as for the xLSTM-TS. The main differences are the Batch size, which is set at 256, and the Sequence Length. There was indeed no significant improvements when increasing the Sequence Length from 100 to 150 so it was fixed at 100 to reduce the computation time.

\begin{table}[htbp]
\caption{Optimal Hyper-parameter Configuration for Darts Models}
\begin{center}
\begin{tabular}{|c|c|}
\hline
\textbf{Parameter} & \textbf{Value} \\
\hline
Learning Rate & 0.0001 \\
Batch Size & 256 \\
Sequence Length & 100 \\
Number of Epochs & 200 \\
Patience & 10 \\
Gradient Clipping Max Norm & 1.0 \\
Optimiser & Adam \\
Loss Function & MSE \\
\hline
\end{tabular}
\label{table_hyperparameters_darts}
\end{center}
\end{table}

\subsection{Evaluation Metrics}

In order to evaluate the effectiveness of predictive models, we need some objective measures. Evaluation metrics such as Accuracy, F1 Score or MAE each give an idea of the performance of the models. In this paper, we used the following metrics:

\begin{itemize}
  \item \textbf{Accuracy}: Measures the proportion of true results among the total number of cases examined.
  
  \item \textbf{Recall}: Indicates the proportion of actual positives that were correctly identified by the model.
  
  \item \textbf{Precision}: Indicates the proportion of positive identifications that were actually correct.
  
  \item \textbf{F1 Score}: Harmonic mean of precision and recall, providing a single score that balances both measures. It is particularly useful when you have uneven class distribution.
  
  \item \textbf{MAE (Mean Absolute Error)}: Measures the average magnitude of the errors in a set of predictions, without considering their direction.
  
  \item \textbf{RMSE (Root Mean Squared Error)}: Measures the square root of the average of the squares of the errors, providing a way to penalise larger errors more heavily than smaller ones.
  
  \item \textbf{RMSSE (Root Mean Squared Scaled Error)}: A scale-independent measure comparing the forecast accuracy to that of a naive forecast method, which predicts the next value as the previous observation \citep{MAKRIDAKIS20221346}.
  
  \item \textbf{MASE (Mean Absolute Scaled Error)}: A scale-independent measure that compares the forecast accuracy to the in-sample MAE from a naive forecast method, which predicts the next value as the previous observation \citep{HYNDMAN2006679}.
\end{itemize}

\begin{table}[htbp]
\caption{Evaluation metrics and their formulas.}
\begin{center}
\renewcommand{\arraystretch}{2.0}
\begin{tabular}{|c|c|}
\hline
\textbf{Evaluation metrics} & \textbf{Formulas} \\
\hline
Accuracy & $\frac{TP + TN}{TP + FP + FN + TN}$ \\
Recall & $\frac{TP}{TP + FN}$ \\
Precision (Positive predictive value) & $\frac{TP}{TP + FP}$ \\
Precision (Negative predictive value) & $\frac{TN}{FN + TN}$ \\
F1 Score & $\frac{2 \cdot \text{Precision} \cdot \text{Recall}}{\text{Precision} + \text{Recall}}$ \\
Mean Absolute Error (MAE) & $\frac{1}{n} \sum_{i=1}^{n} |y_i - \hat{y}_i|$ \\
Root Mean Squared Error (RMSE) & $\sqrt{\frac{1}{n} \sum_{i=1}^{n} (y_i - \hat{y}_i)^2}$ \\
Root Mean Squared Scaled Error (RMSSE) & $\sqrt{\frac{\frac{1}{h} \sum_{i=n+1}^{n+h} (y_i - \hat{y}_i)^2}{\frac{1}{n-1} \sum_{i=2}^{n} (y_i - y_{i-1})^2}}$ \\
Mean Absolute Scaled Error (MASE) & $\frac{\frac{1}{h} \sum_{i=n+1}^{n+h} |y_i - \hat{y}_i|}{\frac{1}{n-1} \sum_{i=2}^{n} |y_i - y_{i-1}|}$ \\
\hline
\end{tabular}
\renewcommand{\arraystretch}{1.0}
\label{table_evaluation_metrics}
\end{center}
\end{table}

\section{Results}

\subsection{Prediction Performance Metrics}

In this section, we present the quantitative results of our predictions using the models introduced earlier. The metrics considered include MAE, RMSE, RMSSE, and MASE. The models were trained with denoised data and tested on the original datasets to evaluate their performance. The predicted values were compared with the actual test set values, and the performance metrics are shown in Table \ref{table_performance_metrics_denoised}.

The datasets used in this study are on different scales, which makes direct comparison of MAE and RMSE values challenging. However, the RMSSE and MASE metrics allow for a standardised comparison of the performance of the models, regardless of the scale of the data.

\begin{table}[htbp]
\caption{Performance Prediction Metrics}
\begin{center}
\begin{scriptsize} 
\begin{tabular}{|c|c|c|c|c|c|}
\hline
\textbf{Dataset} & \textbf{Model} & \textbf{MAE} & \textbf{RMSE} & \textbf{RMSSE} & \textbf{MASE} \\
\hline
\multirow{6}{*}{EWZ Daily} 
& TCN & 0.29 & 0.37 & 0.82 & 0.80 \\
 & N-BEATS & 0.37 & 0.46 & 1.02 & 1.04 \\
 & TFT & 0.57 & 0.74 & 1.63 & 1.59 \\
 & N-HiTS & 0.41 & 0.52 & 1.14 & 1.14 \\
 & TiDE & 0.29 & 0.36 & 0.80 & 0.81 \\
 & \textbf{xLSTM-TS} & \textbf{0.23} & \textbf{0.30} & \textbf{0.58} & \textbf{0.58} \\
\hline
\multirow{6}{*}{S\&P 500 Daily} 
 & TCN & 30.01 & 36.51 & 1.01 & 1.06 \\
 & N-BEATS & 32.14 & 40.82 & 1.13 & 1.13 \\
 & TFT & 48.31 & 64.48 & 1.78 & 1.70 \\
 & N-HiTS & 33.96 & 42.84 & 1.18 & 1.20 \\
 & TiDE & 23.31 & \textbf{29.41} & 0.81 & 0.82 \\
 & \textbf{xLSTM-TS} & \textbf{22.93} & 29.94 & \textbf{0.69} & \textbf{0.70} \\
\hline
\multirow{6}{*}{EWZ Hourly} 
 & TCN & 0.09 & 0.14 & 0.84 & 0.87 \\
 & N-BEATS & 0.15 & 0.20 & 1.24 & 1.48 \\
 & TFT & 0.25 & 0.40 & 2.45 & 2.49 \\
 & N-HiTS & 0.18 & 0.24 & 1.46 & 1.78 \\
 & TiDE & 0.10 & 0.14 & 0.86 & 0.98 \\
 & \textbf{xLSTM-TS} & \textbf{0.06} & \textbf{0.08} & \textbf{0.51} & \textbf{0.65} \\
\hline
\multirow{6}{*}{S\&P 500 Hourly}
 & TCN & 0.95 & 1.35 & 0.96 & 1.05 \\
 & N-BEATS & 0.97 & 1.35 & 0.96 & 1.08 \\
 & TFT & 1.68 & 2.38 & 1.69 & 1.86 \\
 & N-HiTS & 1.21 & 1.65 & 1.17 & 1.34 \\
 & \textbf{TiDE} & \textbf{0.84} & \textbf{1.18} & \textbf{0.83} & \textbf{0.93} \\
 & xLSTM-TS & 1.12 & 1.34 & 0.97 & 1.25 \\
\hline
\end{tabular}
\end{scriptsize}
\label{table_performance_metrics_denoised}
\end{center}
\end{table}

The results demonstrate that the xLSTM-TS model exhibited superior performance across the multiple datasets. For the EWZ daily close prices, xLSTM-TS achieved the lowest MAE of 0.23 and the lowest RMSE of 0.30, indicating the smallest average deviation between predicted and actual values. The model also obtained an RMSSE of 0.58 and a MASE of 0.58, significantly surpassing the naive forecast, which uses the previous observation as the predicted value. Similarly, the xLSTM-TS model achieved better results than the other models on the S\&P 500 daily data, with an MAE of 22.93, an RMSE of 29.94, an RMSSE of 0.69, and a MASE of 0.70, highlighting its robustness.

For the hourly datasets, the xLSTM-TS model continued to show strong performance. On the EWZ hourly dataset, xLSTM-TS achieved the lowest MAE of 0.06, an RMSE of 0.08, an RMSSE of 0.51, and a MASE of 0.65. Although the TiDE model performed well on the S\&P 500 hourly data with an MAE of 0.84, an RMSE of 1.18, an RMSSE of 0.83, and a MASE of 0.93, the xLSTM-TS model's performance remained competitive with an MAE of 1.12, an RMSE of 1.34, an RMSSE of 0.97, and a MASE of 1.25.

It is important to note that a MASE value less than 1 indicates that the model performs better than the naive forecast. The xLSTM-TS, TCN, and TiDE models achieved a MASE below 1 in most cases, indicating effective performance improvements over the naive approach. However, models like N-BEATS and TFT struggled, with MASE values exceeding 1, suggesting their predictions were less accurate than simply using the previous observation.

Furthermore, the difference in prediction accuracy between daily and hourly datasets is notable. Generally, predicting daily close prices results in larger absolute errors compared to hourly prices due to the higher volatility observed over longer time periods. Despite this, the xLSTM-TS model maintained strong performance across both timeframes, demonstrating its adaptability and robustness.

The inclusion of the wavelet denoising process was particularly significant. Prior to denoising, the models' performance on the original, noisy data was close to random guessing, with accuracy metrics around 50\%. The wavelet denoising process effectively enhanced the models' prediction accuracy across various datasets, demonstrating its critical role in improving performance.

\subsection{Directional Movement Prediction Metrics}

This section presents the results for directional movement prediction metrics, focusing on Train Accuracy, Validation Accuracy, Test Accuracy, Precision (Rise), Precision (Fall), and F1 Score. It is important to highlight that the models were trained using denoised data but were evaluated on the original datasets. The performance metrics are detailed in Table \ref{table_directional_metrics_denoised}.

\begin{table*}[htbp]
\caption{Directional Movement Prediction Metrics}
\begin{center}
\begin{tabular}{|c|c|c|c|c|c|c|c|c|}
\hline
\textbf{Dataset} & \textbf{Model} & \textbf{Train Accuracy} & \textbf{Val Accuracy} & \textbf{Test Accuracy} & \textbf{Recall} & \textbf{Precision (Rise)} & \textbf{Precision (Fall)} & \textbf{F1 Score} \\
\hline
\multirow{6}{*}{EWZ Daily} 
 & TCN & 64.21\% & 62.18\% & 66.30\% & 67.63\% & 66.20\% & 66.42\% & 66.90\% \\
 & N-BEATS & 60.21\% & 62.55\% & 62.32\% & 64.75\% & 62.07\% & 62.60\% & 63.38\% \\
 & TFT & 51.87\% & 56.36\% & 52.17\% & 53.24\% & 52.48\% & 51.85\% & 52.86\% \\
 & N-HiTS & 56.80\% & 61.82\% & 60.87\% & 60.43\% & 61.31\% & 60.43\% & 60.87\% \\
 & TiDE & \textbf{69.18\%} & 64.36\% & 69.57\% & \textbf{73.38\%} & 68.46\% & 70.87\% & 70.83\% \\
 & \textbf{xLSTM-TS} & 68.51\% & \textbf{65.33\%} & \textbf{72.87\%} & 72.02\% & \textbf{74.33\%} & \textbf{71.43\%} & \textbf{73.16\%} \\
\hline
\multirow{6}{*}{S\&P 500 Daily} 
 & TCN & 64.23\% & 65.82\% & 63.41\% & 69.86\% & 64.15\% & 62.39\% & 66.89\% \\
 & N-BEATS & 58.56\% & 63.64\% & 59.78\% & 62.33\% & 61.90\% & 57.36\% & 62.12\% \\
 & TFT & 51.90\% & 60.36\% & 53.99\% & 55.48\% & 56.64\% & 51.13\% & 56.06\% \\
 & N-HiTS & 56.38\% & 64.36\% & 56.52\% & 56.85\% & 59.29\% & 53.68\% & 58.04\% \\
 & TiDE & \textbf{67.55\%} & 68.36\% & 64.86\% & 67.81\% & 66.44\% & 62.99\% & 67.12\% \\
 & \textbf{xLSTM-TS} & 67.39\% & \textbf{70.93\%} & \textbf{71.28\%} & \textbf{76.84\%} & \textbf{69.52\%} & \textbf{73.49\%} & \textbf{73.00\%} \\
\hline
\multirow{6}{*}{EWZ Hourly} 
 & TCN & 62.56\% & \textbf{63.48\%} & 62.35\% & 61.65\% & 60.61\% & 64.00\% & 61.13\% \\
 & N-BEATS & 59.31\% & 57.58\% & 55.25\% & 54.26\% & 53.35\% & 57.07\% & 53.80\% \\
 & TFT & 53.64\% & 51.69\% & 51.71\% & 52.27\% & 49.73\% & 53.72\% & 50.97\% \\
 & N-HiTS & 55.26\% & 54.64\% & 52.52\% & 51.99\% & 50.55\% & 54.45\% & 51.26\% \\
 & \textbf{TiDE} & \textbf{65.82\%} & 62.00\% & \textbf{64.26\%} & \textbf{65.91\%} & \textbf{62.03\%} & 66.57\% & \textbf{63.91\%} \\
 & xLSTM-TS & 64.81\% & 62.64\% & 63.63\% & 65.82\% & 60.42\% & \textbf{67.00\%} & 63.00\% \\
\hline
\multirow{6}{*}{S\&P 500 Hourly} 
& TCN & 64.15\% & 60.82\% & 62.35\% & 67.45\% & 62.84\% & 61.73\% & 65.06\% \\
& N-BEATS & 63.24\% & 59.20\% & 60.03\% & 64.04\% & 61.00\% & 58.86\% & 62.48\% \\
& TFT & 55.90\% & 53.90\% & 54.57\% & 57.48\% & 56.15\% & 52.77\% & 56.81\% \\
& N-HiTS & 58.19\% & 57.88\% & 56.34\% & 58.79\% & 57.88\% & 54.62\% & 58.33\% \\
& TiDE & \textbf{66.92\%} & 63.48\% & 63.85\% & 67.72\% & 64.50\% & 63.06\% & 66.07\% \\
& \textbf{xLSTM-TS} & 65.44\% & \textbf{64.57\%} & \textbf{67.23\%} & \textbf{71.53\%} & \textbf{67.32\%} & \textbf{67.11\%} & \textbf{69.36\%} \\
\hline
\end{tabular}
\label{table_directional_metrics_denoised}
\end{center}
\end{table*}

For the daily datasets, the xLSTM-TS model demonstrated outstanding performance. On the EWZ daily data, it achieved a Test Accuracy of 72.87\% and an F1 Score of 73.16\%, surpassing all other models. TiDE also showed strong results with a Test Accuracy of 69.57\% and an F1 Score of 70.83\%. The TCN model followed, while N-BEATS and N-HiTS demonstrated moderate capabilities. TFT struggled significantly in this dataset. In the S\&P 500 daily dataset, xLSTM-TS again led the performance metrics with a Test Accuracy of 71.28\% and an F1 Score of 73.00\%. TiDE was the next best performer, followed by TCN. N-BEATS and N-HiTS showed lower performance, and TFT continued to lag.

In the hourly datasets, TiDE emerged as the top performer for the EWZ hourly data, with a Test Accuracy of 64.26\% and an F1 Score of 63.91\%. The xLSTM-TS model closely followed, achieving a Test Accuracy of 63.63\% and an F1 Score of 63.00\%. TCN also performed well, whereas N-BEATS and N-HiTS provided moderate performance. TFT remained the lowest performer in this dataset. In the S\&P 500 hourly dataset, xLSTM-TS led the results with a Test Accuracy of 67.23\% and an F1 Score of 69.36\%. TiDE followed with strong results, while TCN showed competitive performance. N-BEATS and N-HiTS had moderate outcomes, and once again, TFT recorded the lowest performance.

In summary, the proposed xLSTM-TS model consistently exhibited superior performance across the various datasets. TiDE, a newer approach, also delivered strong results, especially in the hourly datasets for EWZ and S\&P 500. TCN showed competitive performance but was generally outperformed by the previous two. Models such as N-BEATS and N-HiTS provided moderate performance, while TFT consistently underperformed across all datasets. These results underscore the robustness and efficacy of the xLSTM-TS model, highlighting its potential for accurately predicting directional movements in stock prices when trained with denoised data and evaluated on the original data.

\section{Discussion}

\subsection{Model Performance Comparison}

The results highlight the significant improvement in prediction accuracy achieved through the denoising process, particularly for the xLSTM-TS model, which consistently demonstrated superior performance across various datasets.

A closer look at the MASE metric reveals that only the xLSTM-TS, TiDE, and TCN models outperform the naive forecast, which simply uses the previous observation as the predicted value. This underscores the critical importance of selecting and configuring the right model architectures. For example, TFT's performance was not better than random guessing, indicating its unsuitability for this application.

In terms of test accuracy and F1 score, predicting daily close prices proved easier than hourly prices. The xLSTM-TS model achieved F1 scores of around 73\% for both the EWZ and S\&P 500 daily datasets. However, the performance declined for hourly datasets, with F1 scores falling below 70\%. This reflects the greater difficulty in predicting hourly movements, which are subject to higher volatility and smaller price variations.

The EWZ, known for its higher volatility, presented additional challenges. While daily predictions were comparable to the S\&P 500, the hourly predictions for EWZ resulted in a significant drop in F1 scores to 63.91\%. This is notably lower compared to the S\&P 500 hourly F1 score of 69.36\%, highlighting the impact of volatility and time frame on model performance.

TiDE, a newer state-of-the-art model, showed strong results but was ultimately outperformed by the proposed xLSTM-TS model. Interestingly, the older TCN model also performed well, surpassing newer models like N-HiTS. This suggests that well-established models can still offer competitive performance in stock market predictions.

Examining the Train Accuracy and Test Accuracy, particularly for the xLSTM-TS model, we observe an increase in test accuracy, which indicates effective use of configurations like early stopping to prevent overfitting. Overall, the models did not exhibit significant differences between train and test accuracy, suggesting no major fitting issues.

The imbalanced nature of time series data in upward and downward trends necessitates the use of precision and recall metrics. Despite this imbalance, the models did not show extreme differences in performance between movements, indicating balanced prediction capabilities.

It is crucial to note that training models on original, noisy data resulted in performance metrics close to random guessing. The implementation of wavelet denoising was vital, enabling the achievement of high F1 scores up to 73.16\%, a significant improvement given the complexities of stock market prediction.

The RMSSE metric, being scale-independent, allowed for effective performance comparison across all datasets. The consistent performance across different datasets is encouraging, indicating the robustness of the selected models in various stock market environments.

\begin{figure}[htbp]
\centerline{\includegraphics[width=\columnwidth]{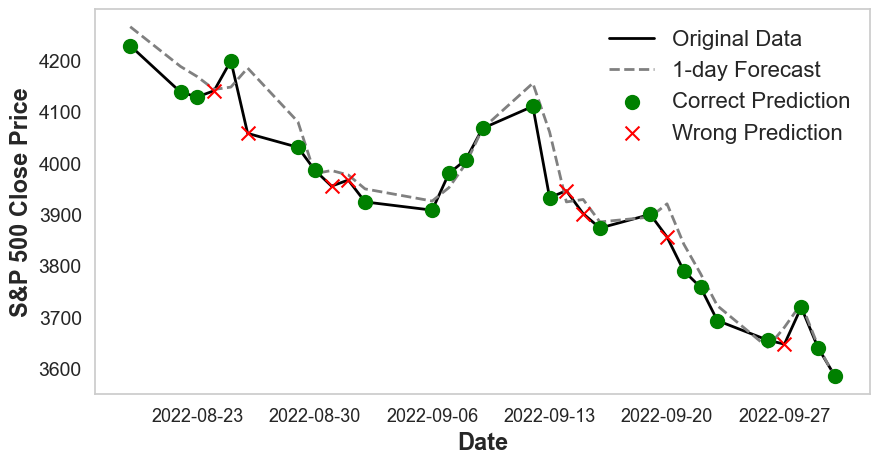}}
\caption{xLSTM-TS next day forecasting showing directional predictions.}
\label{fig-denoised-xLSTM}
\end{figure}

Figure \ref{fig-denoised-xLSTM} illustrates the xLSTM-TS model's next-day predictions compared with the actual S\&P 500 close prices. The figure uses markers to highlight when the model's directional predictions (up or down) were correct (circles) or incorrect (crosses). This visualisation showcases the model's ability to accurately track price trends.

\subsection{Strenghts and Limitations}

The primary strength of this study lies in the successful application of advanced deep learning models, particularly the novel xLSTM adaptation for time series. This model consistently demonstrated superior performance across various datasets, highlighting its robustness and efficacy in predicting stock market trends.

The wavelet denoising process played a crucial role in enhancing model accuracy. By transforming noisy data into a cleaner form, the models could achieve higher prediction accuracy, as demonstrated by the improved F1 scores and other performance metrics. This preprocessing step was instrumental in overcoming the limitations posed by raw, noisy data, which had previously resulted in performance metrics no better than random guessing.

However, several limitations warrant attention. Firstly, while the denoising process significantly improved model performance, relying on these predictions for trading strategies can still lead to incorrect assumptions about stock direction. The models, although trained with denoised data, were evaluated against the actual data, ensuring that the predictions were aligned with real market conditions. Despite this, the inherent risk in following model predictions for buy, hold, or sell decisions remains, as models may not capture sudden market changes or anomalies.

Secondly, the variability in model performance across different datasets and timeframes indicates that the efficacy of these models can be context-dependent. While the xLSTM-TS and TiDE models demonstrated robustness, their performance varied notably between daily and hourly datasets, as well as between the S\&P 500 and EWZ indices.

Moreover, the inherent imbalance in time series data, with more frequent upward trends compared to downward trends, poses a challenge for maintaining balanced prediction capabilities. Although the models showed no extreme differences in performance between movements, the imbalance could affect the generalisability of the results.

In conclusion, while the advanced models and preprocessing techniques employed in this study have shown promising results, careful consideration must be given to the limitations and context-specific performance variations.

\section{Conclusion}

\subsection{Summary of Findings}

This study provided an in-depth analysis of advanced deep learning models for predicting stock market trends, focusing on the S\&P 500 index and the Brazilian ETF EWZ. The findings indicate that the proposed xLSTM-TS model consistently outperformed other state-of-the-art models, demonstrating superior predictive accuracy across multiple datasets. Specifically, the xLSTM-TS achieved the highest Test Accuracy and F1 Score, highlighting its robustness and ability to handle the complexities of financial time series. The use of wavelet denoising significantly enhanced model performance by reducing noise, underscoring the importance of effective preprocessing.

Comparative analysis showed that while the xLSTM-TS model led in performance, the TiDE model also delivered strong results, particularly in hourly datasets. The TCN model demonstrated competitive performance, often surpassing newer models like N-HiTS.

The study found variability in model performance across different datasets and timeframes. Predictions of daily closing prices generally achieved higher accuracy than hourly prices, reflecting the challenges of higher frequency data. Additionally, the volatility of the EWZ index posed greater prediction challenges compared to the S\&P 500.

Evaluation metrics provided a comprehensive assessment of model performance. The RMSSE and MASE metrics were particularly valuable for scale-independent comparisons across datasets, and the F1 Score highlighted the directional accuracy. Overall, the findings highlight the potential of advanced deep learning models, particularly the xLSTM-TS, in improving stock market movement predictions, with denoising playing a critical role.

\subsection{Future Work}

Future research should evaluate these models across a wider range of financial instruments, such as commodities, foreign exchange, and other stock indices, to understand the generalisability of the models' predictive capabilities. Enhancing model robustness in extreme market conditions and integrating additional preprocessing techniques or alternative model architectures would be beneficial.

Applying these models in real-world trading environments and developing automated trading strategies based on their predictions could provide practical insights. Incorporating additional features, such as macroeconomic indicators and sentiment analysis, could further improve predictive accuracy. Future work should focus on feature engineering and selection to enhance model inputs.

In conclusion, this study demonstrated the potential of advanced deep learning models, especially when combined with effective data preprocessing techniques, in improving stock market trend predictions. Further research is needed to address limitations and explore new opportunities for enhancing financial forecasting models.

\begingroup
\small
\bibliographystyle{agsm}
\bibliography{references}
\endgroup

\end{document}